\documentclass[lettersize,journal]{IEEEtran}
\usepackage{amsmath,amsfonts}
\usepackage{algorithmic}
\usepackage{algorithm}
\usepackage{array}
\usepackage[caption=false,font=normalsize,labelfont=sf,textfont=sf]{subfig}
\usepackage{textcomp}
\usepackage{stfloats}
\usepackage{url}
\usepackage{verbatim}
\usepackage{graphicx}
\usepackage{cite}
\usepackage{multirow}
\usepackage{float} 
\begin{document}

\title{Stochastic Decision-Making Framework for Human-Robot Collaboration in Industrial Applications}
\author{Muhammad Adel Yusuf,
        Ali Nasir,
        Zeashan Hameed Khan,%
\thanks{Muhammad Adel Yusuf is with the Department of Control and Instrumentation Engineering, King Fahd University of Petroleum and Minerals (KFUPM), Dhahran 31261, Saudi Arabia (e-mail: Muhammad.adel@ejust.edu.eg).}%
\thanks{Ali Nasir is with the Department of Control and Instrumentation Engineering, and the Interdisciplinary Research Center for Intelligent Manufacturing \& Robotics, King Fahd University of Petroleum and Minerals (KFUPM), Dhahran 31261, Saudi Arabia (e-mail: ali.nasir@kfupm.edu.sa).}%
\thanks{Zeashan Hameed Khan is with the Interdisciplinary Research Center for Intelligent Manufacturing \& Robotics, King Fahd University of Petroleum and Minerals (KFUPM), Dhahran 31261, Saudi Arabia (e-mail: Zeashan.khan@kfupm.edu.sa).}%
}

\markboth{Journal of \LaTeX\ Class Files,~Vol.~14, No.~8, August~2021}%
{Shell \MakeLowercase{\textit{et al.}}: A Sample Article Using IEEEtran.cls for IEEE Journals}

\IEEEpubid{0000--0000/00\$00.00~\copyright~2021 IEEE}

\maketitle

\begin{abstract}
Collaborative robots, or cobots, are increasingly integrated into various industrial and service settings to work efficiently and safely alongside humans. However, for effective human-robot collaboration, robots must reason based on human factors such as motivation level and aggression level. This paper proposes an approach for decision-making in human-robot collaborative (HRC) environments utilizing stochastic modeling. By leveraging probabilistic models and control strategies, the proposed method aims to anticipate human actions and emotions, enabling cobots to adapt their behavior accordingly. So far, most of the research has been done to detect the intentions of human co-workers. This paper discusses the theoretical framework, implementation strategies, simulation results, and potential applications of the bilateral collaboration approach for safety and efficiency in collaborative robotics.
\end{abstract}

\begin{IEEEkeywords}
Cobots, MDP , POMDP, Human Robot collaboration, Industerial robots
\end{IEEEkeywords}

\section{Introduction}
\IEEEPARstart{T}{he} pressing need of modern industries to cope with the technological standards set by Industry 4.0 has resulted in significant advancements in robotics. These developments have facilitated the transition of robots from solitary, safety-constrained environments to cooperative ones where they operate alongside humans. This transition towards collaboration has led to the emergence of collaborative robots (cobots), which are found to be very helpful in executing many complicated tasks even through direct physical interaction or contactless collaboration with human  ~\cite{guda2023introduction}. This collaboration is crucial for achieving the maximum advantages from the human and robot by leveraging the unique capabilities of each. For instance, humans have demonstrated their ability to deal with the dynamic and variable factors of the workplace, while robots excel at doing repetitive jobs with great precision and prolonged endurance. However, there are safety issues that need to be taken into consideration when both humans and the robot are closely integrated in a shared workspace. To ensure safety collaboration between the robot and human in the working environment, these cobots are designed with safety features/components such as force sensors and emergency stop mechanisms to prevent accidents ~\cite{vysocky2016human}. However, in open and unconstrained environments, these robots, alongside the humans need to be integrated with additional sensors (such as visual and non-visual sensors) to monitor their surroundings and detect the presence and actions of human workers, hence guaranteeing smooth and safe collaboration  ~\cite{zacharaki2021decision}. The integration of these components formulates a collaborative ecosystem that, when combined with human interactions, has the potential to trigger hazardous events, thereby posing risks that can make HRC unsafe under certain conditions ~\cite{munoz2018ergonomics}.
Therefore, the emergence of this type of collaborative environment raises the need for safety procedures to be thoroughly reevaluated, considering both the most recent technology developments and the human element—which is crucial in Industry 5.0 ~\cite{alves2023industry}. In this context, it has been noticed that humans use subtle and non-verbal cues, such as eye movements and head gestures, to communicate their intentions while simultaneously interpreting the cues from others to anticipate their actions ~\cite{huang2015using}. This ability allows a group of people working in the same environment to interpret each other, which fosters their collaboration in shared activities. For instance, when two individuals are working together on the same task, one of them may look at an object or make a head gesture to indicate that he is going to reach for it. Without using words, the other person recognizes this cue and modifies their behavior, possibly moving aside or getting ready to help. This type of intention prediction occurs constantly in daily life and is necessary for productive teamwork. Similarly, predicting human behavior is also crucial in HRC, where the human seeks to accomplish a predefined task with the assistance of the robot ~\cite{bauer2008human}. For example, in collaborative industrial assembly jobs, the robot must understand the human's intents, predict his future actions, and anticipate his next action in order to enable smooth and collaboration between the robot and human.

\begin{figure}[!t]
\centering
\includegraphics [width=2.5in]{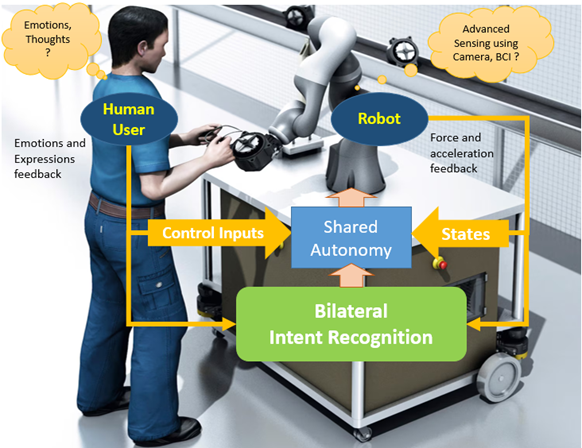}
\caption{Bilateral intent prediction and recognition in Collaboration.}
\label{fig_1}
\end{figure}

In this regard, different sensors have been installed even over the human co-operator or the robot to guarantee effective communication with each other. Among the sensors that have been used in this field to allow the robot and human operator to “see” and communicate with each other are proximity sensors, visual sensors, laser sensors, Inertial sensors, ultrasonic sensors, radar, and acoustic sensors \cite{saleem2024review, liu2023proactive}. Moreover, novel sensors have been designed in some studies that can provide human health state and motion intent in real environments \cite{sawyer2023incorporating}. In similar studies, transparency is assumed to be an important factor for intent detection ~\cite{wang2023effect}. However, there may be uncertainty in the correct prediction -especially with human emotions- that can be regarded as a key performance indicator in robotics tasks such as collision avoidance ~\cite{renz2023uncertainty}. Human comfort is an important factor in the collaborative environment that can ensure a lesser cognitive load while interacting with the machines  ~\cite{yan2023modeling}. Therefore, the need for developing effective intent prediction algorithms based on stochastic modeling and control principles, besides the physical sensors, became a crucial need for collaborative robots to anticipate human emotions and actions more accurately, leading to safer, more efficient, and more intuitive HRC across various domains. Hence, instead of assuming the existence of the human model, this work seeks to adopt a partially observable Markov decision process (POMDP) to learn the human models from the observed data. POMDP offers a broad modelling paradigm for sequential decision-making in which actions have stochastic results and states are hidden. In contrast to most of the existing work on safety-enabled HRC, which considers the musculoskeletal human body model to determine the available workspace for the robot, this work proposes a novel tri-circle approach for safety-related decision-making.

\begin{figure*}[!t]
\centering
\includegraphics {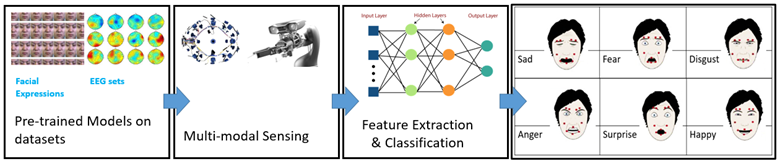}
\caption{Multi-modal sensing and classification of the human mood and well-being.}
\label{fig_2}
\end{figure*}

This work uses stochastic modelling and control approaches to introduce a method that leverages POMDP for HRC environments. The suggested approach seeks to associate the system safety awareness to certain groups of chosen actions, which continuously encourages the HRC system to do the shared task safely and smoothly in a short time and smooth collaboration. The theoretical framework is assessed on a simulated human-robot collaborative scenario and proved capable of identifying loss and success scenarios.
The rest of the paper is organized as follows: Section ~\ref{Section 2} presents the background and reviews previous research for detecting and estimating human Intention. Section ~\ref{Section 3} outlines the theoretical framework of the MDP and POMDP. Section ~\ref{Section 4} introduces the intent prediction algorithm. Section ~\ref{Section 5} details the closed-loop implementation for online trajectory generation. A numerical case study of a human collaborating with a robot in an industry environment is provided in Section ~\ref{Section 6} to validate our proposed approach. Finally, Section ~\ref{Section 7} summarizes the paper with conclusions and our future work.

\section{Background and Related Work}
\label{Section 2}
Intention detection of either the robot or human co-worker represents an essential part, and a critical area of research in HRC, since it allows robots to understand, communicate, and react to human actions and intentions. In this regard, various research studies have been conducted and analyzed from different perspectives in the hope of enhancing the capability of the robot to perceive, interpret, and predict the intentions of its human counterparts, fostering smooth interaction and efficient task execution. One of the most popular techniques in this area is the Vision-Based intent prediction model, which relies on camera systems to capture and analyze human behavior. The camera could be anchored somewhere in the workspace or may be placed on a moving part of the robot. In \cite{khatib2017visual}, Khatib et al. used a depth camera mounted on the worker’s head in order to address the motion control problem between a KUKA robot and a human in a ROS environment. Nevertheless, the detection in this work was not in real-time, and three markers were positioned around the robot for continuous camera localization. On the other hand, there are other works where the vision sensor is mounted on a fixed location such as proposed in \cite{de2012integrated} in which a real time collision avoidance approach has been proposed using a Microsoft Kinect depth sensor mounted on the top of the robot workspace. In the same way, Maric et.al in \cite{maric2023human} have proposed a computationally effective approach that uses RGB-D-based skeleton tracking and hand-crafted modeling to identify human intention in cooperative pick-and-place tasks. However, the popularity of the vision-based approach over their counterparts, it frequently needs a lot of training data and has trouble extrapolating to new situations.

From the HRC researcher’s perspective, coexisting HRC systems have been created to assist human co-operators in controlling industrial robots using simple gestures, voice commands, or even eye movements. Therefore, workers can collaborate with robots to achieve a shared goal without the need for traditional control devices/methods \cite{othman2023human}. In this context, in \cite{lavit2024gaze} Lavit Nicora et al. investigated the adoption of gaze direction to initiate and enhance natural collaboration between the robot and the co-operator in an assembly task. In ~\cite{islam2019understanding} Islam et al. have proposed a simple HRC framework based on hand gesture in order to allow the underwater robots to follow the drivers without the need to translate these gestures into instructions, making it easy for divers to use without needing special tools or memorizing complex rules. Although, these gestures create a seamless collaboration, it suffers from low prediction accuracy in dynamic environments due to noisy data or the existence of obstacles. Neuroscience researchers have also contributed to predict human intentions by establishing a theoretical foundation based on analysing the motion and behaviour of the human body depending on the human brain signals ~\cite{mirabella2014should}. For instance, Horowitz et al. [21] have proposed a feedback mathematical filter to evaluate human arm-reaching intentions in real-time, taking the advantages of the force data collected from the arm to predict the worker's intended reaching trajectory and position. In ~\cite{feleke2021emg}, Electromyography (EMG) signals obtained from human hands have been used to establish a model for predicting the human motor intention. This technique allows the cobot to understand and react to complex hand movements without the need for detailed knowledge of the arm's joint mechanics. This prediction allows the robot to understand and respond to complex hand movements without needing detailed knowledge of the arm's joint mechanics. However, the noise sensitivity of these sensors degrades their efficiency in practical applications.

From the perspective of AI engineers, Intelligent decision using deep learning is found helpful for the gesture recognition of the operator in an industrial environment ~\cite{ding2024smart}. For instance, in \cite{kamali2023vr}, Kamali Mohammadzadeh et al. have developed a virtual platform that facilitates realistic and context-aware human-robot collaboration while ensuring safety. Therefore, an unsupervised machine learning approach that combines dynamic time warping and k-means clustering has been utilized in order to allow the cobots to understand human intentions without requiring labelled data. For the same reason, Lagamtzis et al. in ~\cite{lagamtzis2023exploiting}, have proposed a novel Graph Neural Network (GNN) architecture taking the advantages of the successful application of graph-based methods for recognizing human actions and predicting 3D motion.  In ~\cite{zhang2022reinforcement}, an HRC algorithm has been built based on a reinforcement learning algorithm that has been proposed to optimize work sequence allocation in assembly operations. Other recent studies covered intention detection for social mobile robots that interact with humans very closely for human-aware navigation  ~\cite{alaguero2023communicating}. In the field of industrial robotic applications, ~\cite{cai2024fedhip} Cai et al. describes a federated learning approach to safeguard workers in a factory automation system. While autonomous robots are commonly used in the framework of industry 4.0, it is imperative to apply the fundamental concepts of motion estimation-based intention prediction as found in \cite{chandramowleeswaran2023implementation}. Shared control experiments using neural feedback of the Brain Computer Interface (BCI) are one of the possible use-case for safety approaches leading to improved transparency, performance, and safety ~\cite{dimova2023error}. Wearable devices interfaced with automation system are found to help in safe interfacing with the collaborative environment ~\cite{kahanowich2024learning}. In some cases, intuitive approaches such as power and force limiting are used as a risk mitigation approach as in ~\cite{landmann2023towards}. To verify and compare approaches, various datasets are used in the simulated response tasks in the robotics domain to assess the validity of approaches \cite{orlov2024rw4t, yao2024virtual}. As shown in Figure. \ref{fig_2}, a multi-modal sensing (such as facial expressions using camera and EEG using BCI headset) for the human state of emotion could provide a measure of human well-being to the Robot safety system in order to execute collaborative tasks. Indeed if the human expressions are stressful, it may compromise the safety of both the men and machine within the industrial environment and as a precautionary measure the safety system may halt the Robot from performing any subsequent task.  

Due to the limitations found for the previous HRC approaches and in order to achieve an efficient and safe collaboration between the robot and the operator, researchers have proposed the use of probabilistic models. Among various probabilistic sequential models, such as Markov chains and Hidden Markov Models (HMMs), POMDP model has proven its capability to offer a more generalized approach by taking the partial observability into consideration. This makes it possible for POMDPs to deal with observing uncertainties brought on by concealed human actions and the robot's subpar sensory abilities  ~\cite{broz2011designing, nikolaidis2015efficient}. The fundamental idea of the POMDP approach is revolved around converted the planning and low/high level control problem in HRC into a comprehensive model for sequential decision making under uncertainties  ~\cite{zheng2018pomdp, zhang2019performance}. This model has been found in many applications such as privacy and security ~\cite{ahmadi2018privacy} and robotics ~\cite{chatterjee2015qualitative} and more recently in HRC ~\cite{jean2015pomdp}. Besides POMDP, other probabilistic and stochastic models have been previously presented for discussing the systematic level design problem in HRC.  For instance, in ~\cite{wang2013probabilistic}, an Intention-Driven Dynamics Model (IDDM) in combination with previous work employing Gaussian Mixture Models (GMMs) has been employed to generate robot control policies through the use of reinforcement learning. In \cite{lasota2013developing}, human and robot behaviors are modeled using MDP, which incorporates uncertainties through the construction of control policies based on reward functions. Similarly, a Mixed-Observability MDP (MOMDP), in which states are a combination of partially observable and observable variables, is used in \cite{nikolaidis2015efficient}. In order to represent human intentions and preferences, the reward function is learned by Inverse Reinforcement Learning (IRL), and the control strategy is determined by solving an optimization problem. All in all, since POMDP extends MDP and MOMDP by handling partial observability of all system states and IDDM and GPDM operate in continuous state spaces, POMDP has been adopted as the most suitable and comprehensive approach for representing uncertainties in HRC.

\begin{table*}[htbp]
\caption{Control Approaches for Various Robot Types and Applications}
\centering
\begin{tabular}{|c|c|c|c|}
\hline
\textbf{Type of Robot} & \textbf{Control Approach} & \textbf{Application} & \textbf{Ref} \\
\hline
Assembly Line Robot & Deep Learning & Gesture Recognition & \cite{ding2024smart}, \cite{du2024learning} \\
Robotic Limb & Multi-Modal & Collaborating Sorting & \cite{du2024learning} \\
Virtual Robot & Shared Control using BCI & Pick and Place & \cite{dimova2023error} \\
Social Robot & Human-Aware Navigation & Path Planning & \cite{alaguero2023communicating} \\
Industrial Robot & EEG-Based & Object Handover & \cite{rajabi2023detecting} \\
Industrial Robot & Workspace Optimization & HRC & \cite{tung2024workspace} \\
Industrial Robot & Shared Autonomy (Gaze-Based) & Action Primitive Recognition & \cite{wang2023gaze} \\
Industrial Robot & Digital Twin-Based Control for Human-Robot Symbiosis & Collaborative Assembly & \cite{zhang2024enabling} \\
\hline
\end{tabular}
\label{tab:control_approaches}
\end{table*}

\begin{table}[htbp] 
\caption{AI Algorithms Used in Human-Robot Collaboration}
\centering
\small 
\resizebox{\columnwidth}{!}{ 
\begin{tabular}{|p{3.5cm}|p{4cm}|p{1.5cm}|} 
\hline
\textbf{AI Method} & \textbf{Application} & \textbf{Ref} \\
\hline
Unsupervised Learning & Context-Aware HRC & \cite{kamali2023vr} \\
Graph Neural Networks & Action Recognition + 3D Motion Forecast & \cite{lagamtzis2023exploiting} \\
Convolutional Neural Networks (CNN) & Human-Robot Collaboration & \cite{liu2024human} \\
Visuo-Lingual Transformers & Human-Robot Interaction & \cite{mathur2023proactive} \\
GAN + LSTM Nets & Cooperative Tasks & \cite{mavsar2023gan} \\
Conditional Prediction & Human-Robot Collaboration & \cite{pandya2024towards} \\
Markov Decision/Prediction & Assembly Line & \cite{precup2023recognising, wang2023modeling} \\
Behavior Modeling & Disassembly & \cite{tian2023optimization} \\
Transformer Network & Human-Robot Collaboration & \cite{zhang2024early} \\
Reactive Temporal Logic Planning & Human-Robot Collaboration & \cite{zhou2023local} \\
\hline
\end{tabular}
} 
\label{tab:ai_algorithms}
\end{table}

\section{Key Aspects of the Proposed Approach}
\label{Section 3}
Before we present our mathematical model for decision-making, it is important to highlight the major contribution of this paper by describing the key aspects that are included in the proposed model. Based on the extensive review of literature, we are confident that our proposed approach is unique in terms of combining safety, emotion-based reasoning, and task management in a single decision-making framework. 

\subsection{Safety Circles}
\begin{figure}[!t]
\centering
\includegraphics {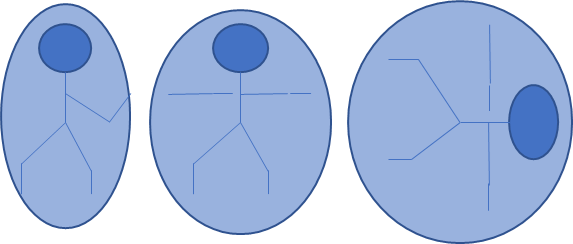}
\caption{Three circles of safety.}
\label{fig_3}
\end{figure}

Most of the existing work on safety-enabled HRC considers the musculoskeletal human body model to determine the available workspace for the robot. In our work, we propose a tri-circle approach for safety-related decision-making. Figure. \ref{fig_3} shows the three circles of safety, where the first circle encapsulates the working space of a human (while the human is engaged in a certain task). This circle is considered by multiple existing approaches. The second circle of safety represents a human with fully stretched arms. This circle represents an aggressive physical act by the human. The third circle represents a human falling over, e.g., in case of an accident. In order to be fully safe, the robot needs to consider the largest circle of safety, but that results in a diminished workspace for the robot. On the other hand, for maximum availability of the workspace, the robot should consider the smallest circle of safety, but then the safety is not guaranteed for the cases of a human stretching his arms unexpectedly (or falling over accidentally). Therefore, the robot must have a decision-making mechanism to decide when to choose which circle of safety. Using semantic information (human pose data) from the Robot’s visual sensors, it is possible to exactly predict the safety strategy on the robot’s end that could guarantee a safe distance a robot could maintain under emergency conditions.

\subsection{Human Emotional States and Motivation Level}
Another factor in HRC is the human mood (or emotional state) and the motivation level of the human. Every day is different for humans because of many things going on in their lives that are related to their personal and professional lives. It is not feasible to know about the root cause of a human emotional behavior or motivational level. However, it is possible to determine the human emotional state and infer the motivation level of a human. The determination of human emotional state and motivation is a significant factor in the human-robot collaboration. Therefore, we propose this feature to be part of our framework. 

\begin{figure*}[!t]
\centering
\includegraphics {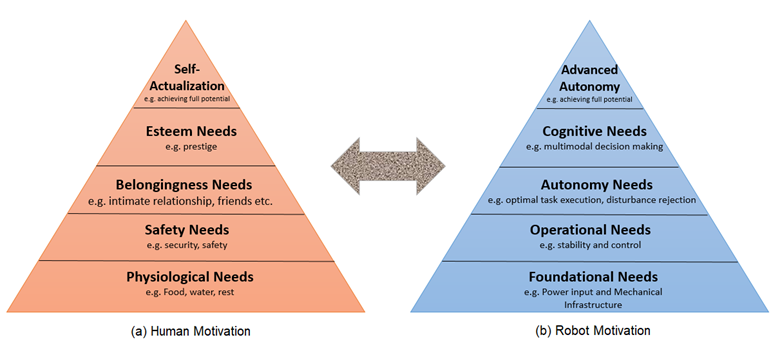}
\caption{An equivalence between Maslow's theory of motivation for humans vs. Robot’s motivation in Collaboration.}
\label{fig_4}
\end{figure*}

In Figure. ~\ref{fig_4}, an equivalence of the Maslow’s hierarchy of needs is elaborated and proposed for the Robot’s motivation. This theory proposes a step-by-step progress to reach the self-actualization to deploy full human potential and, in case of a robot, requires advanced autonomy based on multi-modal sensing and an AI decision support system to play an equal supportive role in decision making.

\subsection{Task Assignment and Priorities}
Another significant factor in the determination of human behavior is the tasks assigned to the human and their corresponding priorities. It is a common observation that human behavior is influenced by the rewards and cost of action choices. The awareness about the cost and rewards available to humans corresponding to various activities is likely to improve the accuracy of human intent prediction. Therefore, this factor is also considered in our formulation.

\section{Human-Robot Collaboration Model}
\label{Section 4}
In this section, we develop POMDP model for collaborative decision-making. As discussed in the preceding section, we must incorporate three factors in the decision-making model, i.e., safety, human emotion, and task assignment. Furthermore, the model needs to be practically solvable as POMDPs are computationally complex in general.

\subsection{State Space}
Keeping all the factors into consideration, we propose the following state space for the model:
\begin{align}
S &= \{s_1, s_2, \dots, s_N\} \\
s_i &= \{\tau_i, d_i, e_i, h_i, b_i\}, \quad i \in \{1,2,3,\dots,N\} \\
\tau_i &= \{\tau_{pi}, \tau{ci}, \tau{xi}, \tau{yi}\} \\
\tau_{pi} &\in \{0,1,2,\dots,\rho\}, \\
\tau{ci} &\in \{0,1,2,3,\dots,\sigma\} \\
\tau{xi}&\in \{0,1,2,3\} \\
\tau{yi} &\in \{0,1,2,3\} \\
d_i &\in \{0,1,2,3\} \\
e_i &= \{e_{mi}, e_{ai}\}, \\
e_{mi}& \in \{0,1,2\} \\
e_{ai}& \in \{0,1,2\} \\
h_i &= \{h_{pi}, h_{ci}\}, \\
h_{pi}&\in \{0,1,2,\dots,\rho\}, \\
h_{ci}& \in \{0,1,2,3,\dots,\sigma\} \\
b_i &= \{b_{pi}, b_{ci}\}, \\
b_{pi} &\in \{0,1,2,\dots,\rho\}, \\
b_{ci}& \in \{0,1,2,3,\dots,\sigma\} \\
N &= (\rho \times \sigma)^3 \times 36
\end{align}

\begin{figure}[!t]
\centering
\includegraphics {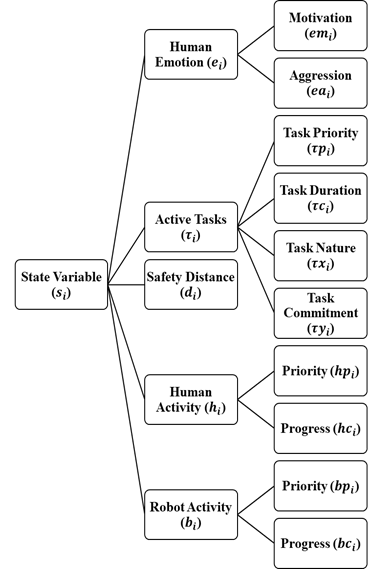}
\caption{State Variable Mapping.}
\label{fig_5}
\end{figure}

We have included five types of variables in the state space, namely, task assigned ($\tau_i$), distance between the robot and the human ($d_i$), emotion of the human ($e_i$), current human activity ($h_i$), and current robot activity ($b_i$). 
The task variable includes information regarding task priority ($\tau_{pi}$), task duration ($\tau_{ci}$), task nature ($\tau_{xi}$), and task commitment status ($\tau_{yi}$). Task priority ranges between zero (lowest priority) to $\rho$ (highest priority). The units of task duration are in terms of the decision epochs ranging between zero (meaning that the task has been completed) and $\sigma$ (meaning that the task will need $\sigma$ decision epochs to complete). The nature of the task ($\tau_{xi}$) identifies whether the task can be accomplished using the robot only ($\tau_{xi}=0$), the human only ($\tau_{xi}=1$), either the robot or the human ($\tau_{xi}=2$) or needs both human and robot ($\tau_{xi}=3$). Finally, the task commitment variable ($\tau_{yi}$) indicates whether the assigned task is uncommitted ($\tau_{yi}=0$), committed by the human ($\tau_{yi}=1$), committed by the robot ($\tau_{yi}=2$), or committed by the human and the robot as a joint venture ($\tau_{yi}=3$). 
The distance variable ($d_i$) is defined in terms of the safety circles defined earlier where the value zero corresponds to the collision, one corresponds to the smallest safety circle distance, and three corresponds to the largest safety circle distance. Any distance greater than the largest safety circle is considered as the largest safety circle distance. 
Next, the human emotion variable ($e_i$) includes the motivation level ($e_{mi}$) that ranges from low motivation ($e_{mi}=0$) to high motivation ($e_{mi}=2$). Another emotion-related variable is human aggression level ($e_{ai}$) that ranges from low ($e_{ai}=0$) to high ($e_{ai}=2$). The purpose of the motivation variable is clear as per our discussion in the preceding section. The aggression variable has been included from the safety distance point of view, e.g., if the human is working aggressively, then it is more desirable for the robot to maintain a larger circle of safety. 
The fourth type of variable in our state space is current human activity ($h_i$), which includes the priority of the activity ($h_{pi}$) and the progress of the activity ($h_{ci}$). Note that the priority variable here is similar to the one associated with the assigned task (ranging between zero and $\rho$). Similarly, the task progress variable ($h_{ci}$) is similar to the task deadline variable ($\tau_{ci}$), i.e., the value of $h_{ci}$ indicates how many more decision epochs are required by the human to complete the current activity. 
The fifth type of variable is the current activity of the robot ($b_i$), which includes the priority of the activity ($b_{pi}$) and the progress of the activity ($b_{ci}$). 
Figure ~\ref{fig_5} shows the map of the state space as discussed above.

\subsection{Decision Variables}
The decision-making in our setup is done by the robot only. The decisions made by the human are treated as exogenous events for the robot. Based on the state space defined above, the decision variables available to the robot include the following:

\begin{equation}
D = \{\text{nw}, \text{rh}_1, \text{rh}_2, \text{ct}, \text{dp}, \text{dm}, \text{mh}, \text{dn} \}
\label{eq:decision_set}
\end{equation}

Note that, unlike the elements in the set of states, the elements in the set of decisions are not variables with real values. Rather, each of these elements corresponds to some action taken by the robot. Specifically, \textbf{nw} refers to “normal work mode,” in which case, the robot shall continue to be in the same state of working (or not working) for the next decision epoch except for $b_c$ and $h_c$. Next is \textbf{rh\textsubscript{1}}, which refers to requesting the human only to take up the assigned task, and \textbf{rh\textsubscript{2}}, which refers to requesting the human to take up the assigned task alongside the robot. \textbf{mh} refers to motivating the human. Decision \textbf{ct} refers to “commit to task,” where the robot itself commits to the assigned task. The decisions \textbf{dp} and \textbf{dm} correspond to increasing and decreasing the distance between the human and the robot, respectively. Finally, the decision \textbf{dn} refers to the “Do nothing” mode, which is to be activated when there is no task to be pursued or in case of collision or a high chance of collision. These actions are summarized and described in Table ~\ref{tab3}.

\begin{table}[htbp]
\caption{Description of the Actions}
\centering
\begin{tabular}{|c|p{6cm}|} 
\hline
\textbf{Notation} & \textbf{Description} \\
\hline
nw  & Normal working \\
rh1 & Request human to do the assigned task alone \\
rh2 & Request human to do the assigned task in collaboration with the robot \\  
ct  & Commit the task \\
dp  & Increase the distance between the human and the robot for safety purposes. \\
dm  & Decrease the distance between the human and the robot when collaboration requires close interaction. \\
mh  & Motivate the human to improve task performance and maintain engagement. \\
dn  & Do nothing, idle mode activated when there is no task to be done. \\
\hline
\end{tabular}
\label{tab3}
\end{table}

\subsection{Observation Variables}

The observation variables included in the proposed model concern the position of the human body, joint angles, and angular velocities for the human joints. Upon observing these variables, the robot can assess the probability distribution for the relevant state variables. Mathematically, the set of observation variables is given by:

\begin{equation}
O = \{ q_0, q_{ls}, q_{rs}, q_{le}, q_{re}, q_{lh}, q_{rh}, \dot{q}_0, \dot{q}_{ls}, \dot{q}_{rs}, \dot{q}_{le}, \dot{q}_{re}, \dot{q}_{lh}, \dot{q}_{rh} \}
\label{eq:observations}
\end{equation}

\begin{equation}
q_*, \dot{q}_* \in \{0,1,2,\dots,y\}
\end{equation}

Equation ~\ref{eq:observations} includes the position of the center of mass for the human body ($q_0$) along with the corresponding velocity ($\dot{q}_0$). Furthermore, we have the joint angles and velocities for the left shoulder ($q_{ls}, \dot{q}_{ls}$), right shoulder ($q_{rs}, \dot{q}_{rs}$), left elbow ($q_{le}, \dot{q}_{le}$), right elbow ($q_{re}, \dot{q}_{re}$), left hand ($q_{lh}, \dot{q}_{lh}$), and right hand ($q_{rh}, \dot{q}_{rh}$). Note that we have discretized the values from zero to $y$ because our framework is designed for high-level decision-making, not for low-level control where precise values of these variables may be required. From the observation of these variables, the robot can infer the probability distribution for the stochastic variables in the state space, such as human motivation level ($e_m$) and human aggression ($e_a$).

\subsection{State Transition Mapping }
There are five types of state variables. One type is formulated as random and unknown (with a known conditional probability distribution), i.e., the human emotion, including the human motivation level ($e_m$) and the human aggression level ($e_a$). Task assignment ($\tau$) is assumed to be an exogenous variable (the variable is known at any instant, but its transition in the next decision epoch depends upon the unmodeled user). The distance between the human and the robot ($d$), as well as the current activities of the human ($h_i$) and the robot ($b_i$), are known. However, the transition of the distance is assumed to be stochastic with a known conditional transition probability distribution. The state transition also depends on the decision taken by the robot. 

Consequently, the state transition mapping is a function of the form:

\begin{equation}
T(s_j, D, s_i), \quad \forall i,j \in \{1,2,\dots,N\}
\end{equation}

with $6N^2$ entries (because the size of the decision space is six, and the size of the state space is $N$). The formulation of the state transition probabilities is expressed as follows:

\begin{equation}
P(s_j \mid s_i, O, D) = P(e_{m_j}, e_{a_j} \mid e_{m_i}, e_{a_i}, O, D)
\label{eq:state_transition}
\end{equation}

To complete the transition model, we also need to know the observation probabilities, given by:

\begin{equation}
\begin{aligned}
    P(O \mid e_{m_j}, e_{a_j}, D) &= P( q_0, q_{ls}, q_{rs}, q_{le}, q_{re}, q_{lh}, q_{rh}, \\
    &\quad \dot{q}_0, \dot{q}_{ls}, \dot{q}_{rs}, \dot{q}_{le}, \dot{q}_{re}, \dot{q}_{lh}, \dot{q}_{rh} \mid e_{m_j}, e_{a_j}, D)
\end{aligned}
\label{eq:observation_probability}
\end{equation}

We would like to point out that the only action of the robot that is likely to impact human emotion is when the robot requests the human to commit to an assigned task, i.e., $D = rh$. Furthermore, the transition for the variable $h_c$ from one decision epoch to the next follows the rule:

\begin{equation}
h_{c_j} =
\begin{cases} 
\tau_{c_i}, & \text{if human commits} \\
\max(0, (h_{c_i} - 1)), & \text{otherwise}
\end{cases}
\label{eq:hc_transition}
\end{equation}

Note that in the above rule, the commitment by the human is an exogenous event. The transition of the variable $b_c$ follows the rule:

\begin{equation}
b_{c_j} =
\begin{cases} 
\tau_{c_i}, & \text{if } D = ct \\
\max(0, (b_{c_i} - 1)), & \text{otherwise}
\end{cases}
\label{eq:bc_transition}
\end{equation}

In the above rule, the main difference from the previous one is that the commitment by the robot is done via the decision-making policy. The variable $h_p$ follows the transition rule:

\begin{equation}
h_{p_j} =
\begin{cases} 
\tau_{p_i}, & \text{if human commits} \\
h_{p_i}, & \text{else if } h_{c_i} \neq 0 \\
0, & \text{otherwise}
\end{cases}
\label{eq:hp_transition}
\end{equation}

Here again, the commitment by the human is an exogenous event. Similarly, the variable $b_p$ follows the transition rule:

\begin{equation}
b_{p_j} =
\begin{cases} 
\tau_{p_i}, & \text{if } D = ct \\
b_{p_i}, & \text{else if } b_{c_i} \neq 0 \\
0, & \text{otherwise}
\end{cases}
\label{eq:bp_transition}
\end{equation}

Here again, the commitment by the robot is made through the decision-making policy. The transition in the task assignment variable is assumed to be exogenous and does not follow any rule or probability distribution in our model. However, once the task is assigned, it can be committed by the robot or the human, resulting in the following transition rule:

\begin{equation}
\tau_{y_j} =
\begin{cases} 
3, & \text{if committed by human and } D = ct \\
2, & \text{else if } D = ct \\
1, & \text{else if committed by human} \\
\tau_{y_i}, & \text{otherwise}
\end{cases}
\label{eq:tauy_transition}
\end{equation}

Note that, in the above transition rule, the commitment by the human is an exogenous event. Finally, the transition in the distance variable follows the rule:

\begin{equation}
d_j =
\begin{cases} 
\min(3, d_i + 1 - \Delta), & \text{if } D = dp \\
\max(1, d_i - 1 - \Delta), & \text{else if } D = dm \\
d_i - \Delta, & \text{otherwise}
\end{cases}
\label{eq:distance_transition}
\end{equation}

In the above rule, $\Delta \in \{0,-1,1\}$ is a random variable that has been included to model the random motion of the human, causing a change in the distance between the robot and the human. The probability distribution of this variable is given by:

\begin{equation}
P(\Delta) = \frac{P(\Delta \mid e_{a_i}) P(e_{a_i})}{P(e_{a_i} \mid \Delta)}
\label{eq:delta_probability}
\end{equation}

This completes our description of the state transitions. Note that two types of variables have randomness involved. This implies that to use our proposed framework, the associated probabilities have to be either calculated, estimated, or learned. We believe that obtaining the required probability distributions must be feasible owing to the current state of the art in machine learning.

\subsection{Formulation of the Cost Function}
The cost function formulation is based on the objectives of the proposed framework. In our case, the objectives are the following:
\begin{enumerate}
\item Avoid collision with the human (keep a safe distance),
\item Complete the assigned task as soon as possible, and
\item Exercise regard for the human emotional state.
\end{enumerate}
Consequently, we formulate the reward function as follows:
\begin{equation}
J(s) = k_1 f_1 (\tau_p, \tau_c, \tau_y, b_c, b_p, h_c, h_p, d) + k_2 f_2 (d, e_a, e_m)
\end{equation}
The above cost function is composed of two terms. The first term corresponds to the task completion objective. The function $f_1$ may have many possible realizations as long as the cost is proportional to the time required to complete the task and commitment by either the human or the robot (or both). The cost of having an uncommitted task is proportional to the task priority. Similarly, the cost of time required to complete the task is the sum of the time required to execute the task and the time required to finish any ongoing activity before the commitment to the task.
The second term in the cost function is related to keeping a safe distance from the human that depends upon the aggression level of the human. The function $f_2$ can be defined in multiple ways as long as the objectives of collision avoidance and having regard for human emotion are satisfied. For example, the recommended safe distance for a higher level of aggression should be more than that for a lower level of aggression. Sample realization (used later in the simulation-based case study) of the functions $f_1$ and $f_2$ is provided as follows in Equations \eqref{eq:f1} and \eqref{eq:f2}:
\begin{equation} \label{eq:f1}
\begin{aligned}
    f_1 &= [\alpha_1 (\tau_y=3) \cdot d \cdot h_c \cdot b_c] \\
        &+ [(\alpha_2 + \alpha_3 (b_p=0) + \alpha_4 (h_p=0)) (\tau_y=0) (\tau_p \neq 0)] \\
        &+ [\alpha_5 (2-e_m)(\tau_x=2)(\tau_y=0)] \\
        &+ [\alpha_6 (\tau_x=3)(\tau_y \neq 3)] \\
        &+ [\alpha_7 (\tau_p > b_p)(\tau_y=0)] \\
        &+ [\alpha_8 b_c] + [\alpha_9 h_c]
\end{aligned}
\end{equation}

\begin{equation} \label{eq:f2}
\begin{aligned}
    f_2 &= [\beta_1 (3-d)] \\
        &+ [(\tau_y \neq 3) (\beta_2 e^{(\beta_3 (e_a-3)d)} \\
        &\quad + \beta_4 e^{(-\beta_5 (e_m+1)d)} \\
        &\quad + e_a \cdot \beta_6 e^{-d} )]
\end{aligned}
\end{equation}

Note that Equation \eqref{eq:f1} is composed of seven terms. The first term penalizes the distance between the human and the robot when they are working in collaboration. This term will compete with the safety cost to ensure efficient collaboration. The second term in Equation \eqref{eq:f2} penalizes unassigned tasks (using $\alpha_2$) and robot or human staying idle despite the availability of a task to be completed. The next term penalizes a low motivation level of the human. The fourth term penalizes the wrong task assignment. The fifth term penalizes the robot for doing a low-priority task despite the availability of a high-priority task. The last two terms penalize the robot and human for an incomplete task.
The sample safety cost function in Equation \eqref{eq:f2} consists of two main terms. The first term penalizes the lack of distance between the human and the robot. The second term penalizes the lack of distance from the perspective of human aggression level ($e_a$) and human motivation level ($e_m$).

\subsection{Decision-Making under Uncertainty and MDP}
\begin{figure}[!t]
\centering
\includegraphics[width=0.8\linewidth]{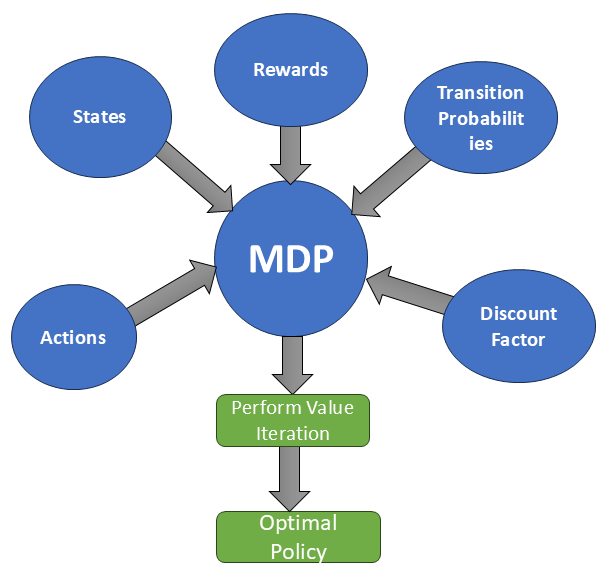}
\caption{Offline policy implementation of MDP.}
\label{fig_6}
\end{figure}

One of the most widely used frameworks for calculating an optimal policy under uncertainty is the MDP. The MDP, as shown in Figure. ~\ref{fig_6}, provides a structured approach to model complex decision-making problems, enabling the calculation of an optimal policy through stochastic dynamic programming techniques, such as policy iteration and value iteration. To mathematically formulate an MDP problem, five key elements must be predefined: a set of states, a set of actions, a reward (or cost) function, a state-action transition probability matrix, and a discount factor within the interval $(0,1)$. An MDP is thus characterized as a five-tuple:

\begin{equation}
MDP = \{S, D, R, P, \gamma\}
\label{eq:MDP_definition}
\end{equation}

where the set of states $S$ represents possible configurations or conditions of the system. The action set $D$ contains all possible actions that can be taken to transition the system from one state toward a desired goal state. The reward function $R$ assigns a value to each state-action pair, distinguishing between favorable and unfavorable outcomes, and can represent the cost or reward of specific actions. The probability tensor $P$ captures the transition dynamics, defined as:

\begin{equation}
P(s' \mid s, d)
\end{equation}

which gives the probability of reaching state $s' \in S$ from state $s \in S$ after performing action $d \in D$. Finally, the discount factor $\gamma$ determines how future rewards are valued relative to immediate ones. A discount factor close to $1$ implies that future rewards are valued almost as highly as current ones, while a lower $\gamma$ places more emphasis on immediate rewards. In addition to the five core elements, an MDP also requires a specified decision-making horizon, which defines the time frame over which decisions are evaluated. This includes determining the decision epoch—the interval between consecutive decision points. In our application, the decision epoch corresponds to the time required for the Cobot to complete its task, which can be identified in seconds, minutes, hours, or even according to the processing unit of the robot. Once a decision-making problem is modeled as an MDP, it can be solved using various stochastic dynamic programming algorithms, such as policy iteration or value iteration. In this paper, we use value iteration, as it allows us to leverage the optimal values of local MDP states when calculating the global decision—a process that will be elaborated upon in the next section. The value iteration algorithm determines the optimal value for each state by applying Bellman’s equation, as shown in Eq.~\eqref{eq:bellman_equation}:

\begin{equation}
V^{(t+1)}(s) = \max_{d \in D} \left\{ -J(s, d) + \sum_{s' \in S} \gamma P(s' \mid s, d) V^t(s') \right\}
\label{eq:bellman_equation}
\end{equation}

To begin the value iteration process, the values for each state are initially set to arbitrary values. After a certain number of iterations, the state values converge, meaning that:

\begin{equation}
V^{(t+1)}(s) = V^t(s), \quad \forall s \in S
\end{equation}

regardless of the initial values. This iterative process continues until a specified stopping criterion is met, typically based on the difference between consecutive iterations, ensuring that the values have stabilized to a satisfactory level of precision:

\begin{equation}
\| V^{(t+1)} - V^t \|_{\infty} < \eta
\label{eq:convergence_condition}
\end{equation}

Here, $V^t$ represents the vector containing the values of all states at time $t$. Once the convergence criterion is satisfied, the resulting state values are considered optimal and are represented as:

\begin{equation}
V^*(s), \quad \forall s \in S.
\end{equation}

These optimal values are then used to derive the optimal policy, calculated according to the following equation:

\begin{equation}
\pi(s) = \arg\max_{d \in D} \left\{ -J(s, d) + \sum_{s' \in S} \gamma P(s' \mid s, d) V^*(s') \right\}
\label{eq:optimal_policy}
\end{equation}

Note that $\pi(s) \in D$, i.e., $\pi: S \to D$.

Due to the existence of true states of the system, such as human emotions, which are not directly observable, the decision-making process becomes extremely uncertain, necessitating the use of POMDP. A POMDP is defined as a combination of an MDP to model system dynamics with a hidden Markov model (HMM) that connects unobservable system states to quantifiable observations~\cite{guda2023introduction}. To extend the previously implemented MDP framework to a POMDP, the present model is improved by adding observation and belief states to account for hidden/uncertain states that cannot be directly measured. By adding an observation space and an observation model, the POMDP expands on the MDP structure. The observation model describes the likelihood of obtaining a specific observation given the system's present hidden state, whereas the observation space contains quantifiable signals that indirectly reflect the hidden states. 

\begin{figure}[!t]
\centering
\includegraphics[width=0.8\linewidth]{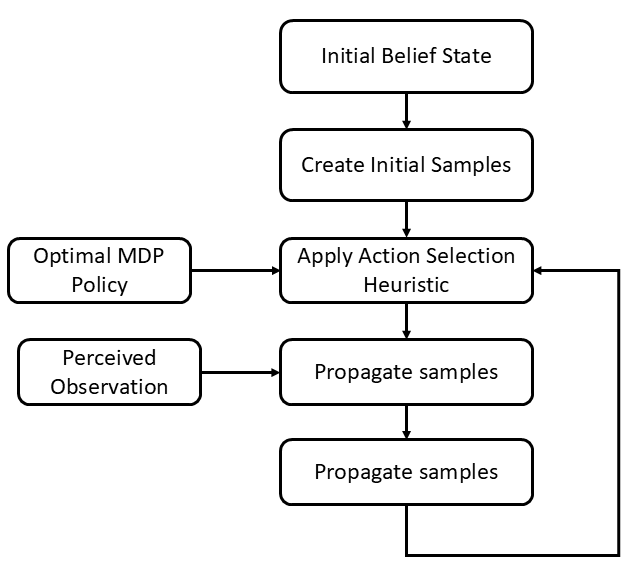}
\caption{Framework of executing the POMDP.}
\label{fig_7}
\end{figure}

As shown in Figure.~\ref{fig_7}, the implementation begins with an initial belief state, which represents a probability distribution over possible hidden states. From this belief, initial samples are created to approximate the system's state space.  Using the optimal MDP policy, obtained previously through MDP, an action is selected using a probabilistic or heuristic technique that considers the score calculation of each action based on the belief state.
The chosen action is then applied, and the samples are propagated through the system's transition model to predict the new state distribution.  After that, an observation is sent to the system, which indirectly reveals the actual state. This observation is incorporated into the belief state, improving the probability distribution over all potential states. By adjusting to the dynamics of the system and the unpredictability of its hidden states, the updated belief state then feeds back into the action selection process, forming an iterative loop that continuously enhances decision-making.

\begin{figure}[!t]
\centering
\includegraphics [scale=0.5]{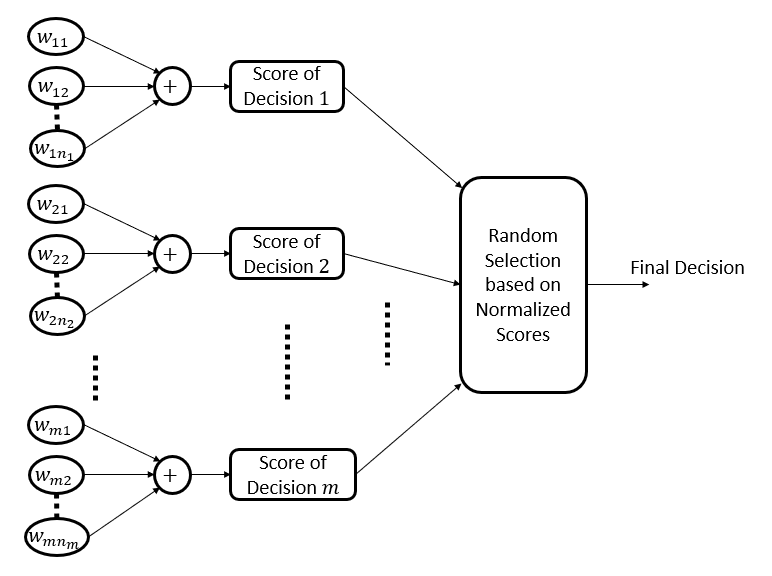}
\caption{Action Selection Heuristic.}
\label{ASH}
\end{figure}

Our proposed action selection heuristic is presented in Figure.~\ref{ASH} In this heuristic, $w_{ij}$ represents the weight of $j^{th}$ sample (state) that gives $i^{th}$ decision as an optimal one as per the policy calculated from the MDP. To calculate the best decision at any instant, the weights are added for each corresponding decision, and whichever decision has the highest sum of weights is declared as the best decision, and the same is executed for all samples.

Furthermore, in our proposed action selection heuristic, we exploit the structure of the optimal policy by gathering all the states that have the same optimal decision as per the pre-calculated MDP policy. On top of that, since the weights are proportional to the probabilities of the states being the actual state, we add the weights to determine the best action. Consider the example of a robot in a maze. Assuming that the states in which turning left is the optimal action are highly likely as per the current belief state, then our action selection heuristic would choose turning left as the best action. If, for some reason, two or more actions have an equal score, then multiple ideas can be used to break the ties between the actions. For example, the highest individual weight value can be used as a tie breaker, e.g., the decision that can an individual weight whose value is larger than all individual weights of the other decision shall win.

By integrating the POMDP framework into the existing MDP implementation, the system can make intelligent and flexible decisions in the face of ambiguity. This allows the system to account for the dynamic and hidden nature of true states, such as human emotions. This method improves the model's resilience and suitability for real-world situations where it is impractical to directly observe crucial states.

\section{Closed loop implementation (Online trajectory)}
\label{Section 5}
\begin{figure}[!t]
\centering
\includegraphics[width=0.8\linewidth]{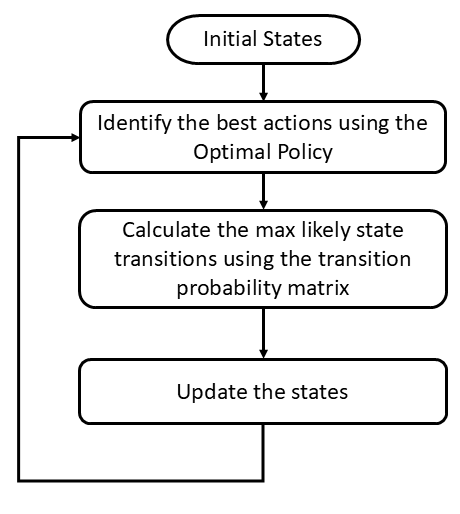}
\caption{online execution of optimal policy.}
\label{fig_8}
\end{figure}

After determining the optimal policy through the value iteration technique, the robot’s online trajectory control begins, as illustrated in Figure. ~\ref{fig_8} This process initiates by setting the initial states of the system.  In the subsequent step, the optimal actions are identified by referencing the pre-computed optimal policy, which ensures that the chosen actions maximize the expected reward. Once the best actions are selected, the likely state transitions are computed using the transition probability tensor, which captures the probability of moving from one state to another based on the selected actions. This step is critical for accounting for uncertainty in the environment and for predicting the robot's next state with high accuracy. Finally, the states are updated according to the identified transitions, and the loop is completed by returning to the second step, where the policy is re-evaluated in light of the new states. This closed-loop process enables the robot to adapt its actions dynamically, optimizing its trajectory in real time based on the evolving states and environmental conditions.

\section{Numerical Case study }
\label{Section 6}
In this section, we perform a detailed analysis of the proposed framework with the help of a numerical simulation-based case study. In our case study, we aim to find the optimal policy that can ensure performing the required task in a very short time taking into consideration the safe collaboration between the robot and the human. The next subsection provides detailed parameter values, followed by an analysis of the closed-loop execution results and key trends observed in the optimal task assignment policy derived through the proposed framework.

\subsection{Parameter values}
There are three types of parameter values: cost function parameters, transition probability parameters, and problem size parameters. Table ~\ref{tab4} details all parameter values and the definitions of the abstract functions. These values were selected roughly to estimate the relation between the time needed for the robot to accomplish the task and ensuring safe collaboration with the human. However, they are intended solely for demonstration and should not be applied directly to real-life missions without further validation

\begin{table}[htbp]
\caption{Parameter Values for the Case Study}
\centering
\begin{tabular}{|c|c|p{5.5cm}|}
\hline
\textbf{Parameter} & \textbf{Value} & \textbf{Remark} \\
\hline
$n$  & 419904  & Number of states. \\
$k$  & 20  & Number of iterations. \\
$D$  & 8  & Number of decision variables. \\
$\tau_{p_i}$ & \{0,1,2\} & “0” for no task, “1” for low-priority task, “2” for high-priority task. \\
$\tau_{c_i}$ & \{0,1,2\} & “0” for task completion, “1” the task needs 1 epoch for completion, “2” the task needs 2 epochs for completion. \\
$h_{p_i}$ & \{0,1,2\} & “0” for no human priority, “1” for low human priority, “2” for high human priority. \\
$h_{c_i}$ & \{0,1,2\} & “0” for human completed assigned task, “1” human needs 1 epoch, “2” human needs 2 epochs to complete the assigned task. \\
$b_{p_i}$ & \{0,1,2\} & “0” for no robot priority, “1” for low robot priority, “2” for high robot priority. \\
$b_{c_i}$ & \{0,1,2\} & “0” for robot completed assigned task, “1” robot needs 1 epoch, “2” robot needs 2 epochs to complete the assigned task. \\

\multirow{2}{*}{$\alpha_1$ to $\alpha_{13}$} &  
\multirow{2}{*}{1}  
& Coefficients of cost function. \\
& & (Values are all set to 1) \\

$\beta_1, \beta_2$ & 1, 1 & Coefficients of cost function. \\
$O_i$ & \{1,2,3\} & Observation of human velocity: “1” for low velocity, “2” for medium velocity, “3” for high velocity. \\
\hline
\end{tabular}
\label{tab4}
\end{table}

\subsection{Sample trajectories}
In this subsection, we examine the outcomes of applying the optimal task assignment policy under different initial conditions to demonstrate how a robot can collaborate with a human in shared environments to complete tasks while maintaining human safety. In our case study, three different tasks have been introduced to the POMDP framework with the following characteristics: 

\begin{equation}
S_1 = \begin{bmatrix} 0,0,1,1,3,0,0,0,0,0,1 \end{bmatrix}
\label{S1}
\end{equation}

\begin{equation}
S_2 = \begin{bmatrix} 0,0,2,2,1,0,0,0,0,0,1 \end{bmatrix}
\label{S2}
\end{equation}

\begin{equation}
S_3 = \begin{bmatrix} 0,0,2,2,2,0,0,0,0,0,1 \end{bmatrix}
\label{S3}
\end{equation}

These tasks have been submitted to our framework sequentially each with its own characteristics to evaluate its applicability to execute different tasks under different scenarios efficiently in a collaboration environment while ensuring human safety. The first task, indicated by $S_1$ in \ref{S1}, is characterized by the following features: (1) it is a medium-priority task ($S_1(3) = 1$ or $\tau_p = 1$); (2) it requires one epoch to complete ($S_1(4) = 1$ or $\tau_c = 1$); and (3) it must be executed collaboratively by both the robot and the human ($S_1(5) = 3$ or $\tau_x = 3$). Then, after 7 epochs, another task, defined in Eq. \ref{S2}, is introduced. This task has the following characteristic: $S_1(3) =2$, $S_1(4) =2$, $S_1(5) =1$ which indicates a high priority task ($\tau_p =2$) with time duration of 2 epochs ($\tau_c =2$), it can be executed by human only ($\tau_x =1$). Finally, the third task is introduced at epoch 14 with the following features: $S_1(3) =2$, $S_1(4) =2$, $S_1(5) =2$ which indicates a high priority task ($\tau_p =2$) with time duration of 2 epochs ($\tau_c =2$), it can be executed by human only ($\tau_x =2$). The rest of the states are initially started with "zero values" to represent the initial conditions of the human, the robot, and the task prior to execution. The only exception is the eleventh state, where $S(11) = 1$, indicating that the initial distance between the human and the robot is short. These initial states reflect that the task is initially unassigned ($S(6) =0$), both the human and the robot begin with zero task progress and low priority activity levels ($S(7,8,9,10) =0$), and human emotions are at the low level ($S(1,2) =0$).

Since human emotions ($e_m$ and $e_a$) cannot be directly measured, these states are inferred based on observable variables such as the human body position, joint angles, and angular velocities of the human joints in real-world scenarios. Upon observing these variables, the robot can assess the probability distribution of the relevant state variables. In our simulation experiments, the human velocity in the workspace serves as an observation. Three velocity levels (low, medium, and high) are used to infer the human emotional state. Given that there are two unobservable states, the belief state consists of nine possible states, as outlined in Table ~\ref{tab5}. If the worker's velocity, as determined by the sensors mounted on the worker joints, is small ($O=1$), the system is most likely in State 1 ($e_m=0$ and $e_a=0$), with a probability of 80\%. There is a smaller chance (10\%) that the system is in State 2 ($e_m=0$ and $e_a=1$), or State 4 ($e_m=1$ and $e_a=0$), while the probabilities for State 3 and States 5 through 9 are zero, indicating that these states are not possible given the observation. When the velocity of the worker is observed to be medium ($O=2$), the belief state is updated. This belief state indicates that the system is most likely in State 5 ($e_m=1$ and $e_a=1$), with a probability of 70\%. There is a 10\% chance that the system is in State 2 ($e_m=0$ and $e_a=1$), or State 4 ($e_m=1$ and $e_a=0$), and a 5\% chance it is in State 6 ($e_m=1$ and $e_a=2$), or State 8 ($e_m=2$ and $e_a=1$). The probabilities for other states are zero, meaning these states are not likely given the observation $O=2$. Finally, if the velocity of the worker is high ($O=3$), the belief state is updated to reflect that the system is most likely in State 9 with a probability of 70\%. There is a 10\% chance that the system is in State 6 ($e_m=1$ and $e_a=2$) or State 8 ($e_m=2$ and $e_a=1$), and a 5\% chance it is in State 3 ($e_m=0$ and $e_a=2$), or State 7 ($e_m=2$ and $e_a=0$). The probability of other states has remained zero.

\begin{table}[htbp]
\caption{Belief States}
\centering
\begin{tabular}{|p{3cm}<{\centering}|p{4cm}<{\centering}|} 
\hline
\textbf{State Number} & \textbf{State Value} \\
\hline
State 1 & $\begin{bmatrix} 0 & 0 \end{bmatrix}$ \\
State 2 & $\begin{bmatrix} 0 & 1 \end{bmatrix}$ \\
State 3 & $\begin{bmatrix} 0 & 2 \end{bmatrix}$ \\
State 4 & $\begin{bmatrix} 1 & 0 \end{bmatrix}$ \\
State 5 & $\begin{bmatrix} 1 & 1 \end{bmatrix}$ \\
State 6 & $\begin{bmatrix} 1 & 2 \end{bmatrix}$ \\
State 7 & $\begin{bmatrix} 2 & 0 \end{bmatrix}$ \\
State 8 & $\begin{bmatrix} 2 & 1 \end{bmatrix}$ \\
State 9 & $\begin{bmatrix} 2 & 2 \end{bmatrix}$ \\
\hline
\end{tabular}
\label{tab5}
\end{table}

The results of our experiments are illustrated in the following figures (\ref{fig_9} -\ref{fig_12}). Figure. \ref{fig_9} illustrates the external inputs that define the characteristics of the active task, including task nature, task priority, task duration, and commitment status. As depicted in this figure, three active tasks have been assigned sequentially each with its own features as illustrated earlier. For instance, the first task begins with a medium priority. Upon completion, the task priority returns to zero before the initiation of the second task. Task 2 and task 3, characterized by a high priority, follows a similar pattern: the priority increases from zero to two during its execution and reverts to zero once completed. The same behavior is observed with the task priority; the figure begins with the expected time for each task completion which decreases at each epoch till the task is completed it goes back to zero. Task nature and task commitment figures are related to each other. The task nature tells how the task should be performed - by Robot only ($\tau_x = 0$), human only ($\tau_x = 1$), either one of them ($\tau_x = 2$) or needs a collaboration between the two agent ($\tau_x = 3$)- while the task commitment ($\tau_y$) tells about the current status of the task. Therefore, once the simulation starts, the task is committed directly according to its nature. This is explicitly illustrated in task nature and task commitment graphs. these graphs show that each of the three tasks start with the initial condition (uncommitted task ($\tau_y=0$)) and with time progression it reaches the task nature before going back again to the zero initial states after task completion.

Figure.~\ref{fig_10} illustrates the status of the human and robot activity during executing the tasks. Since the first task is assumed to be executed in a medium priority collaboration environment between the robot and the worker, both Human and robot activity priority has moved from the low-level (initial state) to the medium-level. then, once the task has been completed, their priorities went back to the low-level again. The same happens with the activity progress, which indicates that the progress of human and robot are aligned with each other since they are engaged in a collaborative task. Moreover, they are also aligned with the value of task duration. For task 2, since its task nature tells that it should be executed by human, only the human activity and human progress will be changed to align with the task priority and task duration. Since the third task could be assigned to either human or robot, the task here is only assigned to the robot. consequently, the priority and progress of only the robot activity is changed to align with the ones of the task while the the priority and progress of the human activity went back to the initial states.

Figure.~\ref{fig_11} shows the relationship between distance and collaboration. Initially, the distance between the human and robot is set at $d = 1$, indicating close proximity, which is maintained during the execution of the first task which requires a high collaboration between the robot and the worker. Upon task completion at epoch 4, the robot re-enters stop mode and the distance changes to $d = 3$. In this mode, since the robot is not operating, the distance is flexible, ranging from collision ($d = 0$) to maximum separation ($d = 3$) without impacting human safety. However, it is preferable to increase the distance at the end of each task to be used as an initial state for the next task to ensure safety. For task 2 and task 3, the task should be executed by either the robot or human, therefore, the distance between the two agents should be high to ensure the safety of the worker and smoothness of the process.

Figure.~\ref{fig_12} shows the sequence of the actions the robot achieved using the optimal policy to achieve its goal. The figure demonstrates how the robot assigned the task to both the robot and human at the same time, keeping the distance between them in the small circle to achieve high collaboration between the robot and the human. The robot starts by motivating the human since the task requires human involvement, and a human with low motivation can impact safety. When motivation reaches a good value (moderately motivated human), the algorithm requests the human to share the task with the robot. Upon completion of the first task, the distance is increased, where the robot and the robot is sent the stop mode. Once the second task starts, which requires the task to be done by human only, the robot starts by motivating the human again to ensure high value of human motivation level. once the robot is motivated, the task is assigned to him. Again, once the task is completed, the robot re-enters the stop mode. Finally, once the third task is assigned to the robot, the robot worked normally until the task is completed after 2 epochs. This process is designed to operate continuously and can theoretically run indefinitely as new tasks are introduced over time. Table~\ref{tab6} summarizes and describes the status of the state variables across the 20 epochs.

\begin{table}[htbp]
\caption{Trajectory Description}
\centering
\small 
\begin{tabular}{|p{1cm}<{\centering}|p{6cm}<{\centering}|} 
\hline
\textbf{Epoch} & \textbf{State Variable Values} \\
\hline
1  & $\begin{bmatrix} 0,0,1,1,3,0,0,0,0,0,1 \end{bmatrix}$ \\
2  & $\begin{bmatrix} 1,1,1,1,3,0,0,0,0,0,1 \end{bmatrix}$ \\
3  & $\begin{bmatrix} 1,1,1,1,3,3,1,1,1,1,1 \end{bmatrix}$ \\
4  & $\begin{bmatrix} 1,1,1,1,3,3,1,0,1,0,2 \end{bmatrix}$ \\
5  & $\begin{bmatrix} 1,1,0,0,0,0,0,0,0,0,3 \end{bmatrix}$ \\
6  & $\begin{bmatrix} 1,1,0,0,0,0,0,0,0,0,3 \end{bmatrix}$ \\
7  & $\begin{bmatrix} 1,1,0,0,0,0,0,0,0,0,3 \end{bmatrix}$ \\
8  & $\begin{bmatrix} 1,1,0,0,0,0,0,0,0,0,3 \end{bmatrix}$ \\
9  & $\begin{bmatrix} 1,1,2,2,1,0,0,0,0,0,3 \end{bmatrix}$ \\
10 & $\begin{bmatrix} 1,1,2,2,1,1,1,2,0,0,3 \end{bmatrix}$ \\
11 & $\begin{bmatrix} 1,1,2,2,1,1,1,0,0,0,3 \end{bmatrix}$ \\
12 & $\begin{bmatrix} 1,1,2,2,1,1,1,0,0,0,3 \end{bmatrix}$ \\
13 & $\begin{bmatrix} 1,1,0,0,0,0,0,0,0,0,3 \end{bmatrix}$ \\
14 & $\begin{bmatrix} 1,1,0,0,0,0,0,0,0,0,3 \end{bmatrix}$ \\
15 & $\begin{bmatrix} 1,1,0,0,0,0,0,0,0,0,3 \end{bmatrix}$ \\
16 & $\begin{bmatrix} 1,1,2,2,2,2,0,0,2,2,3 \end{bmatrix}$ \\
17 & $\begin{bmatrix} 1,1,2,2,2,2,0,0,1,1,3 \end{bmatrix}$ \\
18 & $\begin{bmatrix} 1,1,2,2,2,2,0,0,0,0,3 \end{bmatrix}$ \\
19 & $\begin{bmatrix} 1,1,0,0,0,0,0,0,0,0,3 \end{bmatrix}$ \\
20 & $\begin{bmatrix} 1,1,0,0,0,0,0,0,0,0,3 \end{bmatrix}$ \\
\hline
\end{tabular}
\normalsize 
\label{tab6}
\end{table}

\begin{figure*}[!t]
\centering
\includegraphics[width=0.8\linewidth]{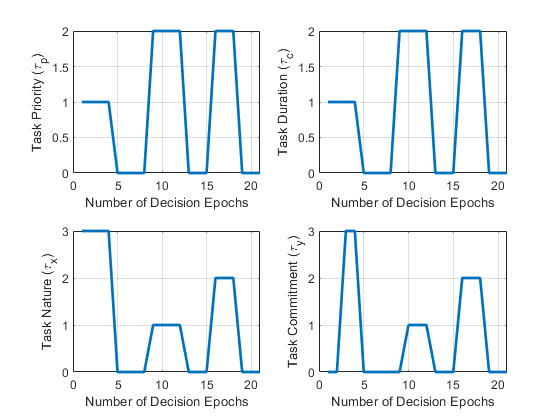}
\caption{Status of The Active Task.}
\label{fig_9}
\end{figure*}

\begin{figure*}[!t]
\centering
\includegraphics[width=0.8\linewidth]{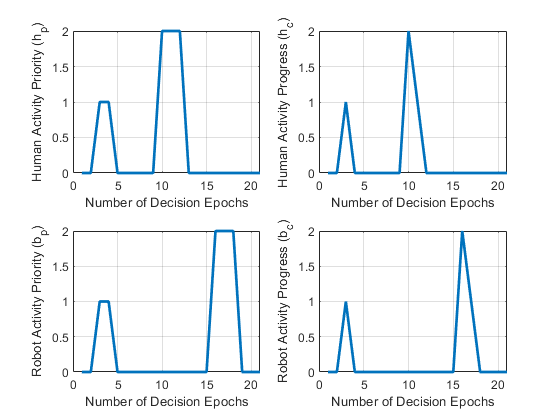}
\caption{Status of The Human and Robot Activity.}
\label{fig_10}
\end{figure*}

\begin{figure}[!t]
\centering
\includegraphics [width=0.8\linewidth]{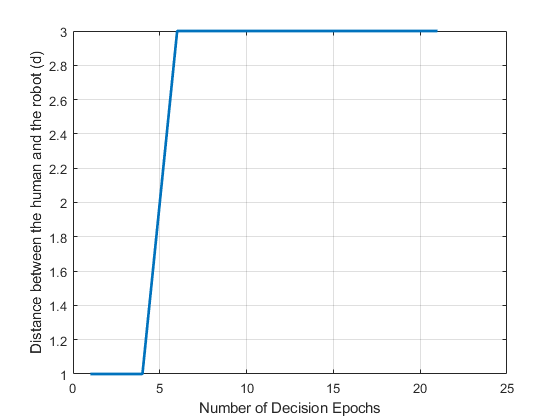}
\caption{Status of the Distance between the Robot and Human.}
\label{fig_11}
\end{figure}

\begin{figure}[!t]
\centering
\includegraphics [width=0.8\linewidth]{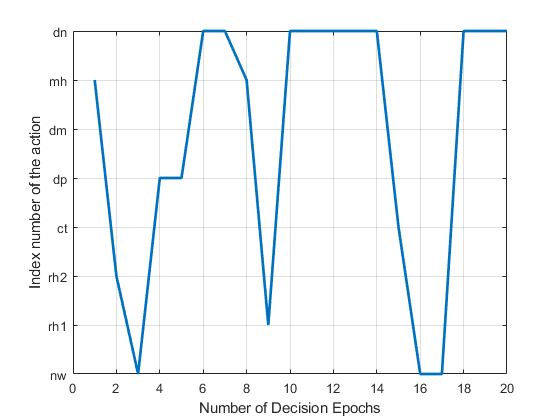}
\caption{Sequence of the actions.}
\label{fig_12}
\end{figure}

\section{Conclusion}
\label{Section 7}
This paper proposes a novel theoretical framework implementation for intent prediction in HRC environments utilizing stochastic modeling and control techniques. By leveraging probabilistic models and a bilateral intent prediction approach, the framework enhances the safety and transparency of collaborative robots, ensuring that they can anticipate human actions and adapt their behavior in real-time. Since most of the research is done either to detect the intention of the robot or the human co-worker, this model excels in considering the bilateral intent prediction approach for safety and transparency in collaborative robotics. Furthermore, the inclusion of both the robot’s state and the uncertain emotional states of the human in a unified hierarchical structure provides a comprehensive solution for dynamic human-robot interactions. Through a numerical simulation-based case study, the paper adeptly demonstrates the practical application of the proposed framework. It not only describes the execution of the framework, but also the consequences of parameter choices.

\bibliographystyle{IEEEtran}
\bibliography{mybibfile}
\vfill
\end{document}